\documentclass[journal,comsoc]{IEEEtran}

\usepackage[T1]{fontenc}

\usepackage{ifpdf}
\usepackage{cite}
\usepackage{graphicx}
\graphicspath{{./Fig/}}
\usepackage{amsmath,amsfonts}
\usepackage{bm}
\usepackage{algorithm}
\usepackage{algorithmic}
\usepackage{array}
\usepackage[caption=false,font=normalsize,labelfont=sf,textfont=sf]{subfig}
\usepackage{url}
\begin{document}

\title{A Preliminary Add-on Differential Drive System for MRI-Compatible Prostate Robotic System}

\author{Zhanyue Zhao, Yiwei Jiang, Charles Bales, Yang Wang, Gregory Fischer
\thanks{Zhanyue Zhao, Yiwei Jiang, Charles Bales, Yang Wang, and Gregory Fischer are with the Department of Robotics Engineering, Worcester Polytechnic Institute, Worcester, MA 01605 USA (e-mail: zzhao4@wpi.edu, gfischer@wpi.edu).}
\thanks{This research is supported by National Institute of Health (NIH) under the National Cancer Institute (NCI) under Grant R01CA166379 and R01EB030539.}
}
\maketitle

\begin{abstract}

MRI-targeted biopsy has shown significant advantages over conventional random sextant biopsy, detecting more clinically significant cancers and improving risk stratification. However, needle targeting accuracy, especially in transperineal MRI-guided biopsies, presents a challenge due to needle deflection. This can negatively impact patient outcomes, leading to repeated sampling and inaccurate diagnoses if cancerous tissue isn't properly collected. To address this, we developed a novel differential drive prototype designed to improve needle control and targeting precision. This system, featuring a 2-degree-of-freedom (2-DOF) MRI-compatible cooperative needle driver, distances the robot from the MRI imaging area, minimizing image artifacts and distortions. By using two motors for simultaneous needle insertion and rotation without relative movement, the design reduces MRI interference. In this work, we introduced two mechanical differential drive designs: the ball screw/spline and lead screw/bushing types, and explored both hollow-type and side-pulley differentials. Validation through low-resolution rapid-prototyping demonstrated the feasibility of differential drives in prostate biopsies, with the custom hollow-type hybrid ultrasonic motor (USM) achieving a rotary speed of 75 rpm. The side-pulley differential further increased the speed to 168 rpm, ideal for needle rotation applications. Accuracy assessments showed minimal errors in both insertion and rotation motions, indicating that this proof-of-concept design holds great promise for further development. Ultimately, the differential drive offers a promising solution to the critical issue of needle targeting accuracy in MRI-guided prostate biopsies.

\end{abstract}


\section{Introduction}

\IEEEPARstart{P}{rostate} cancer is the most common cancer among men in the US. MRI-targeted biopsy has emerged as a superior approach for diagnosing prostate cancer, surpassing the conventional random sextant biopsy. This method has proven to detect more clinically significant cancers than traditional random biopsies, aiding physicians in better risk stratification of prostate cancer. While the benefits of MRI-targeted biopsy, whether in-bore or through fusion techniques, are becoming increasingly evident, needle targeting accuracy remains a critical challenge, especially in the transperineal approach. Researchers have reported that needle deflection, occurring in both in-bore and fusion biopsies, compromises patient outcomes by necessitating repeated sampling of the biopsy site. This unintended deflection also leads to inaccurate diagnoses if the biopsy fails to collect useful cancerous tissue \cite{chatterjee2019revisiting}.

To address the need for improved needle control and targeting accuracy in image-guided biopsies, researchers are developing cooperatively controlled robots with closed-loop systems that track needles in interventional images and actively compensate for deflection. A 2-DOF MRI-compatible cooperative needle driver with both rotation and insertion functions was previously developed and validated \cite{wartenberg2018towards}. However, given the nature of prostate biopsies, the robot must be positioned very close to the patient, within the MRI’s strong magnetic field. This close proximity can cause image artifacts and geometric distortions, as studied in \cite{carvalho2020demonstration}. Moreover, motor vibrations caused by rapidly changing magnetic gradients and eddy currents can further degrade image quality.

The new differential drive prototype developed in this work addresses these challenges by distancing the robot from the MRI imaging area, leveraging the differential drive operation method. Instead of relying on a rotation mechanism that follows insertion, this design utilizes the combined motion of two motors to achieve both insertion and rotation while remaining fixed to the robot, with no relative movement. As long as the needle and shaft are long enough to reach the target, the robot and motion system can be positioned away from the MRI imaging area, significantly reducing interference with scans compared to the current design.

\section{Differential Principle and Method}

\begin{figure}
    \centering
    \includegraphics[width=0.8\linewidth]{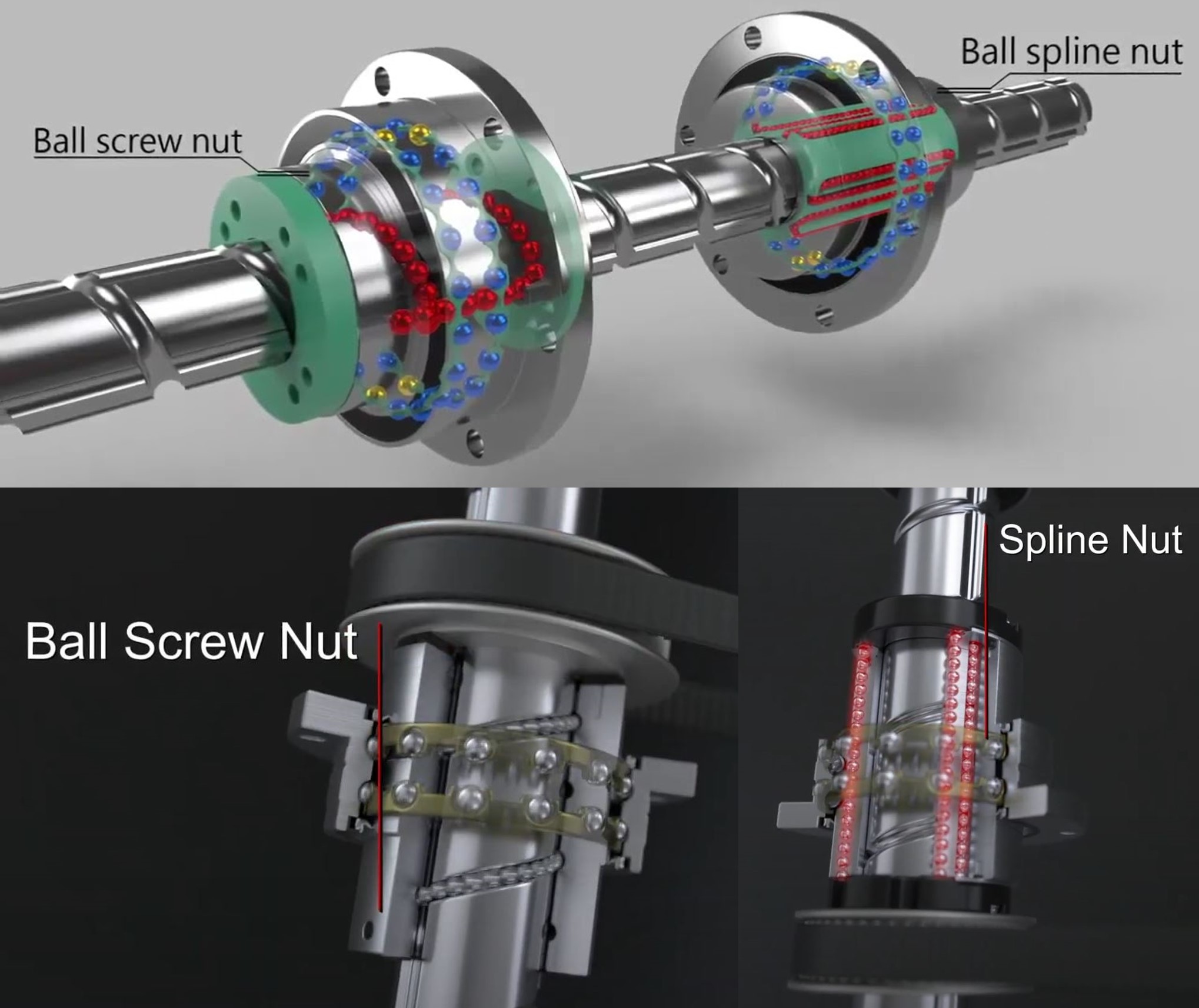}
    \caption{Main components of the ball screw/spline system, consist of a ball screw nut, a ball spline nut, and a shaft with screw and spline grooves. Image from HIWIN\textregistered. Ball screw nut and ball spline nut details. Image from THK\textregistered.}
    \label{fig:BSSsample1}
\end{figure}

The differential drive system is inspired by the ball screw/spline mechanism. The Ball Screw/Spline is a linear-rotary unit that contains Ball Screw grooves and Ball Spline grooves crossing with each other on a single shaft. The nuts of the Ball Screw and the Ball Spline have dedicated support bearings directly embedded in the circumference of the nuts. Figure \ref{fig:BSSsample1} shows the main components of the ball screw/spline system. The nuts of the Ball Screw and the Ball Spline have dedicated support bearings directly embedded in the circumference of the nuts. 

\begin{figure}
    \centering
    \includegraphics[width=0.8\linewidth]{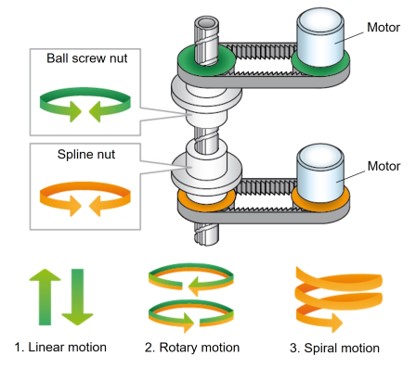}
    \caption{Rotary, linear, and spiral modes of motion with a single shaft by rotating or stopping the ball screw nut or the spline nut. Image from THK\textregistered.}
    \label{fig:BSSsample2}
\end{figure}

The motion modes can be found in Figure \ref{fig:BSSsample2} right figure. In mode 1, the ball screw nut rotates and the spline nut stops, shaft performs linear motion. In mode 2, the ball screw nut and spline nut rotate in the same direction and same speed, shaft performs rotary motion. In mode 3, the ball screw nut stops and the ball screw nut rotates, shaft performs a spiral motion.

The advantages of using a ball screw/spline are significant. First, with zero axial clearance, the Ball Spline has an angular contact structure that causes no backlash in the rotational direction, enabling highly accurate positioning. Second, a lightweight and compact system, since the ball screw nut is integrated with the support bearing, a highly accurate and compact design is allowed. In addition, small inertia through the lightweight ball screw nut ensures high responsiveness. Third, smooth motion with low noise, as the Ball Screw is based on an end-cap mechanism, smooth motion with low noise is achieved. Fourth, highly rigid support bearing, the support bearing on the ball screw and spline have designed contact angles respectively, which provides highly rigid shaft support. In addition, a dedicated rubber seal is attached as a standard to prevent the entry of foreign material. Lastly, for easy installation, the ball spline nut is designed so that balls do not fall off even if the spline nut is removed from the shaft, thus making installation easy.

In addition to the ball screw/spline advantages mentioned above, the differential drive also combines all the advantages of the hollow core motor. By using the differential drive as a needle driver, it is more compact in design and smaller assembly compared to the current system. What's more, the cannula and shaft can be longer, and the differential drive system fixed onto the robot can leave more distance from the MRI bore to minimize the effect on imaging and reduce image artifacts to more degrees. The motion can be used in biopsy needle motion, namely needle insertion and rotation. Based on the principle of ball screw/spline motion modes and principle, two types of differential drive systems were developed, which are ball screw/spline (BSS) and Lead Screw/Bushing (LSB) differential drive systems.

\section{Differential Drive Design}

\subsection{USR60 Based Hollow USM Design}

\begin{figure}
    \centering
    \includegraphics[width=1\linewidth]{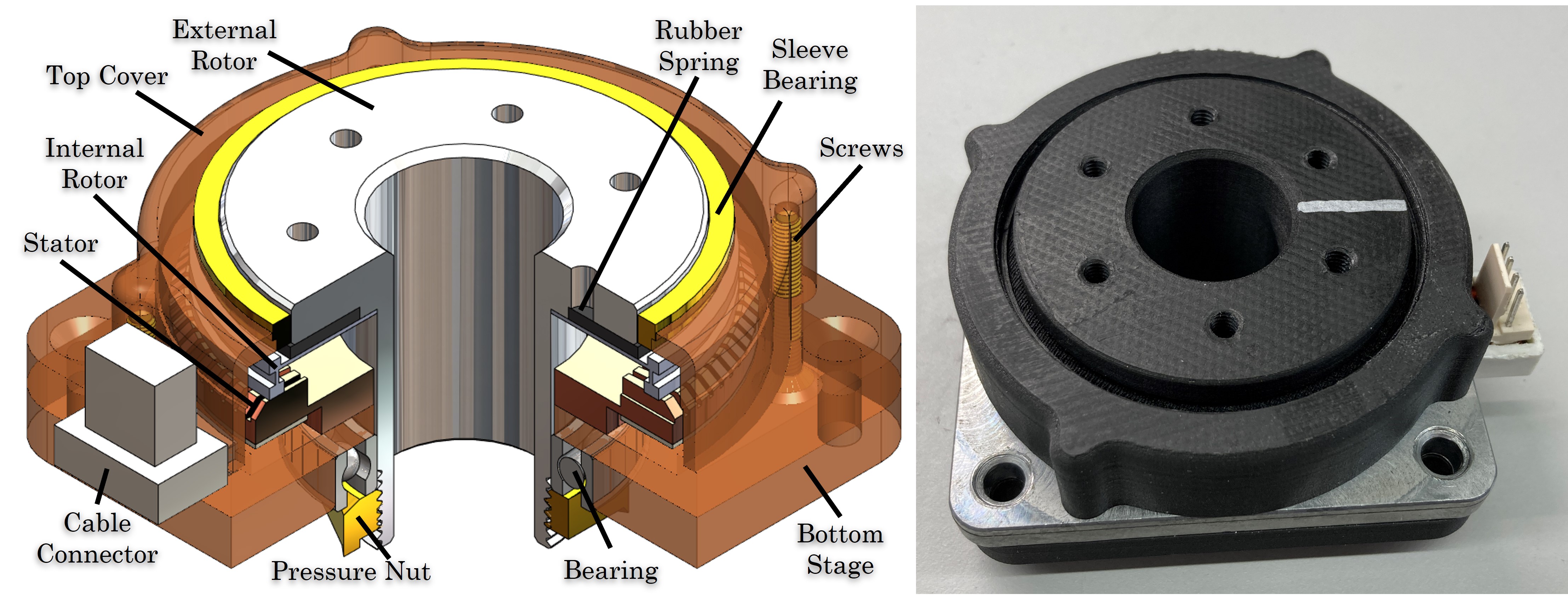}
    \caption{(Left) Design of MRI-compatible hollow shaft motor construction. (Right) Assembly of hollow shaft motor.}
    \label{fig:HSM1}
\end{figure}

\begin{figure}
    \centering
    \includegraphics[width=0.75\linewidth]{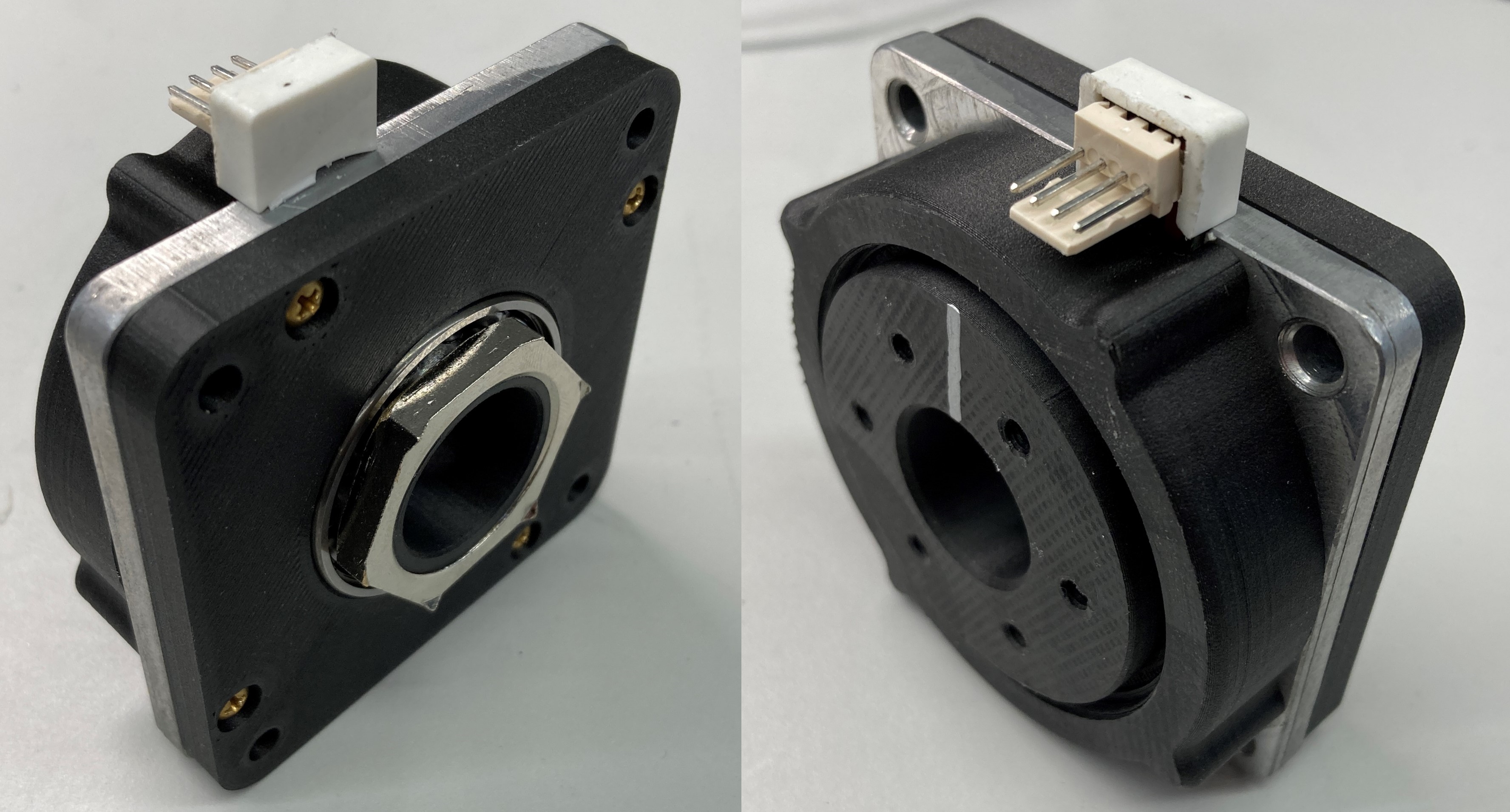}
    \caption{Manufactured hollow shaft USM.}
    \label{fig:HSM2}
\end{figure}

Following our previous work \cite{carvalho2020study,zhao2021preliminary,zhao2023preliminary,zhao2024study,zhao2024design,zhao2024characterization}, a hollow shaft USM (HSM) based on the USR60-NM stator was designed and manufactured in this study. Figure \ref{fig:HSM1} shows the design of MRI-compatible hollow shaft motor construction. The stator was harvested from a Shinsei USR60-NM motor, with a center hole of $\phi$32mm, other components were designed based on this stator. A plastic hollow type rotor, namely an internal rotor ring with a thin surface was covered by a plastic external rotor, where a $\phi$22mm hole was through the center. Between the two rotors, a layer of rubber rings contacted each other and provided both friction and damping space for the external rotor to apply prepressure. An M25$\times$1.5mm thread was manufactured on the bottom tube side of the external rotor. The stator was fixed onto the center hole enlarged original aluminum bottom stage by screws, and a plastic top cover enclosed the stator with rotors assembly, a sleeve bearing was holding the external rotor with the top cover. An additional layer of the plastic bottom stage was attached to the bottom of the stator stage, which was used for holding the bearing that the external rotor went through. A brass M25$\times$1.5mm panel nut matched the external rotor bottom tube thread, which was used to tighten the external rotor and adjust the prepressure applied onto the internal rotor. The top cover and bottom stage were fixed by 4 screws to enclose the whole motor. A manufactured hollow shaft USM can be found in Figure \ref{fig:HSM2}. Note that the motor had to contain the aluminum bottom stage as the first layer for the cable connector to attach onto, this is because the feedback (F.B.) terminal has a filter connected to the GND pin, this filter needs a large contacting aluminum area. Without the contact of the aluminum-made stage, the stator will not work functionally. This needs further research for the F.B. functionality. 

\subsection{Ball Screw/Spline Differential Drive Actuator}

\begin{figure}
    \centering
    \includegraphics[width=1\linewidth]{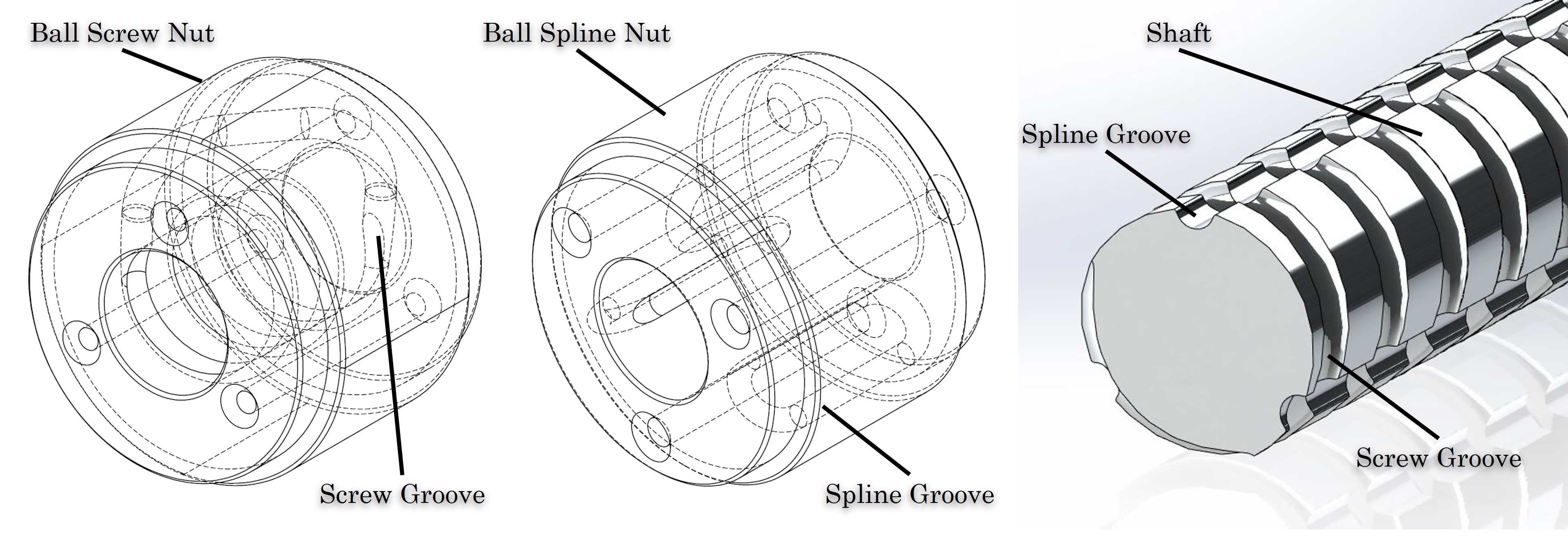}
    \caption{Parts of BSS actuator. (Left) $\phi$20x10mm ball screw nut. (Middle) Ball spline nut with an ID $\phi$20mm center through-hole. (Right) OD $\phi$20mm shaft with both grooves.}
    \label{fig:BSSparts}
\end{figure}

\begin{figure}
    \centering
    \includegraphics[width=0.9\linewidth]{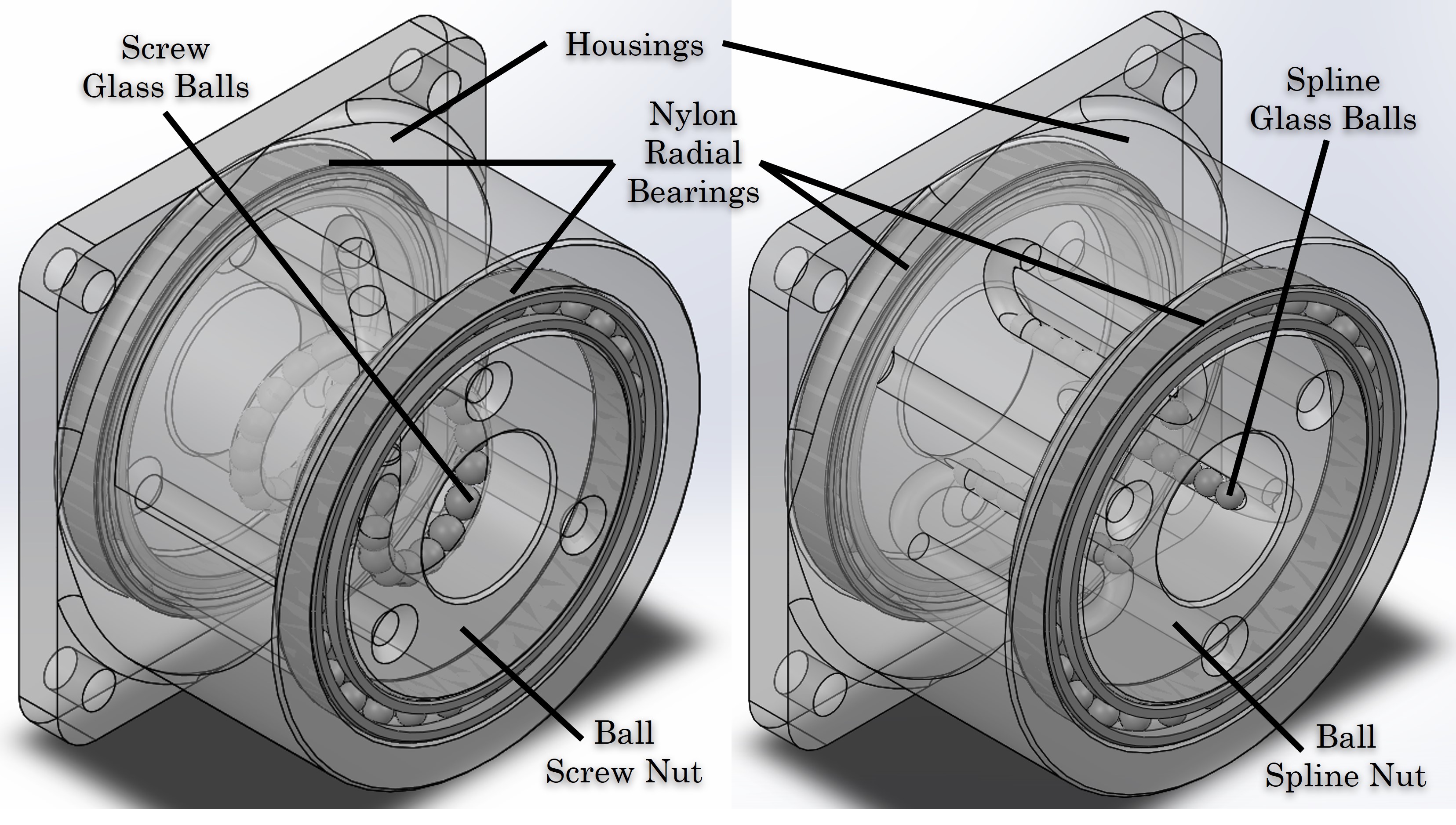}
    \caption{(Left) Ball screw but assembly. (Right) Ball spline nut assembly.}
    \label{fig:bsbs}
\end{figure}

The design drawing of a ball screw/spline (BSS) differential type is similar to the general ball screw/spline system, which consists of a $\phi$20x10mm ball screw nut, a ball spline nut with an ID $\phi$20mm center through-hole, and an OD $\phi$20mm shaft with both grooves. We designed a prototype of the main components, namely the ball screw nut, ball spline nut, and the shaft. Considering the MRI compatibility, we used a Markforged\textregistered~3D printer to print the parts. More specifically, the two nuts were cut in a specific method for easy support removal of the groove tunnel inside the nut body, and the shaft was designed into an identical unit with a 120mm length that can be assembled. The shaft is modular, so it can be extended. Parts of the BSS actuator can be found in Figure \ref{fig:BSSparts}. Figure \ref{fig:bsbs} shows the two nuts assemblies. Ball screw and spline nuts were both held by two 45$\times$58$\times$7mm nylon caged ceramic ball radial bearings (Kashima\textregistered~ Bearing, Japan) UKB6809PP-A, two 3d printed housings were holding both nut-bearings assemblies for smooth rotation. The balls used between the grooves were all 5/32-inch glass balls. 

\begin{figure}
    \centering
    \includegraphics[width=0.9\linewidth]{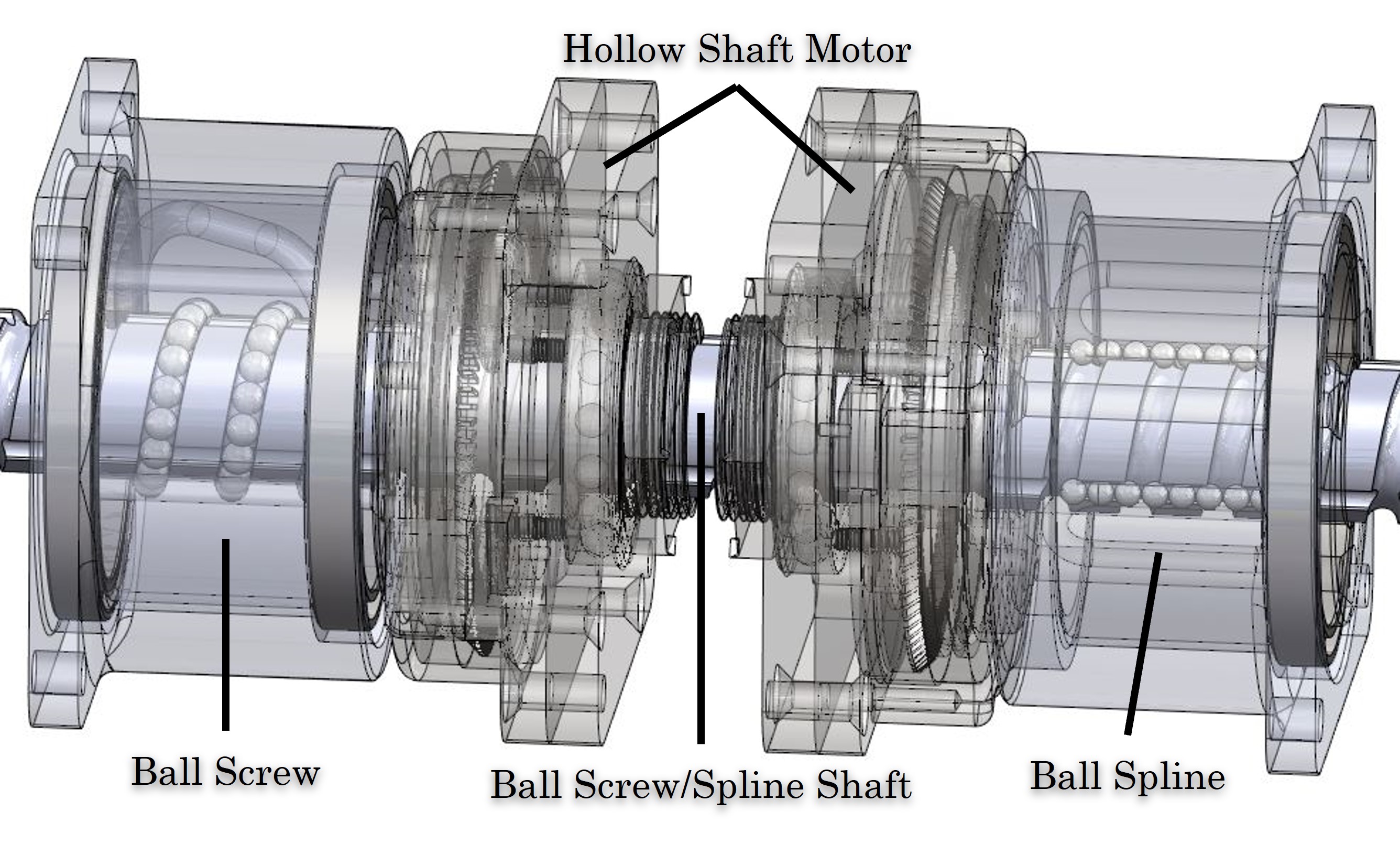}
    \caption{Hollow type of differential drive using ball screw/spline configuration.}
    \label{fig:BSSdesign}
\end{figure}

Figure \ref{fig:BSSdesign} shows the hollow type of differential drive using BSS configuration. Two hollow shaft motors were fixed with a ball screw nut and ball spline nut respectively to drive the rotary motion of nuts, the motor stage motor housings were fixed together relative to the robot, and a grooved shaft was through both nuts and motors contacting with glass ball fitting both grooves. Driving the nuts can perform different modes of motion. Note that the BSS actuator provides smooth motion performance due to the small friction caused by the ball bearings, however, the cost is high and the installation process will be hard with constrain of distance, otherwise the ball will be loose if the shaft is out of the nut bore.

\subsection{Lead Screw/Bushing Differential Drive Actuator}

\begin{figure}
    \centering
    \includegraphics[width=1\linewidth]{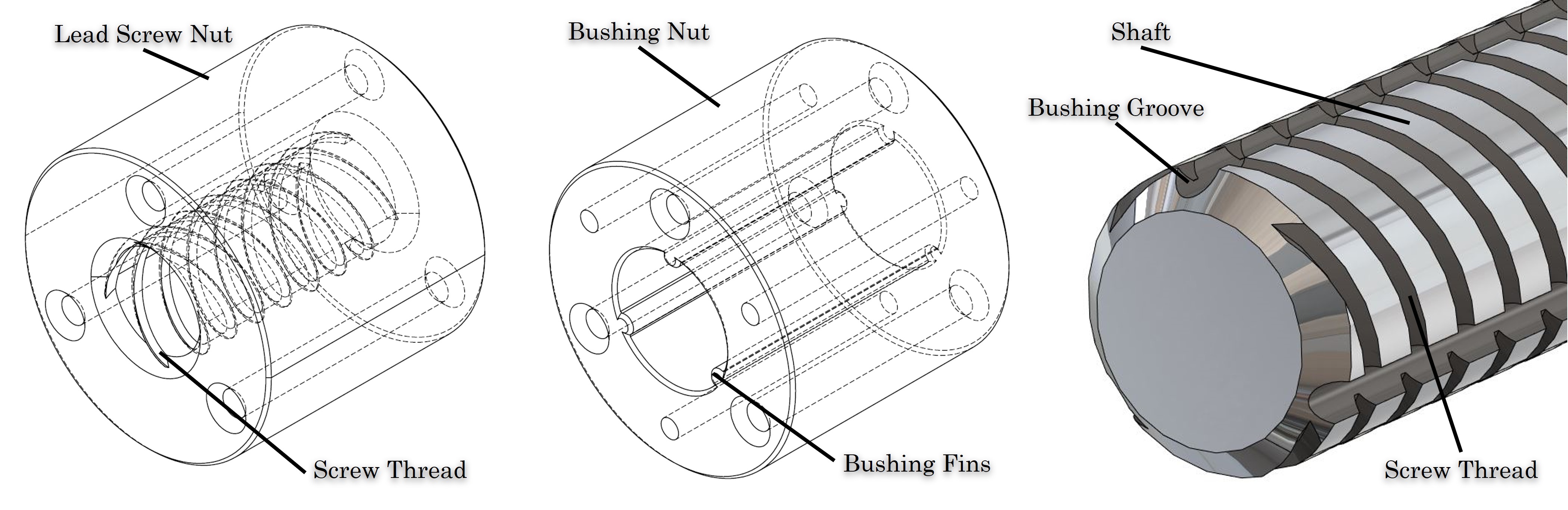}
    \caption{Parts of LSB actuator. (Left) 4-start $\phi$20x20mm lead screw nut. (Middle) Lead bushing nut with an ID $\phi$20mm center through-hole. (Right) OD $\phi$20mm shaft with both bushing groove and screw thread.}
    \label{fig:LSBparts}
\end{figure}

\begin{figure}
    \centering
    \includegraphics[width=0.9\linewidth]{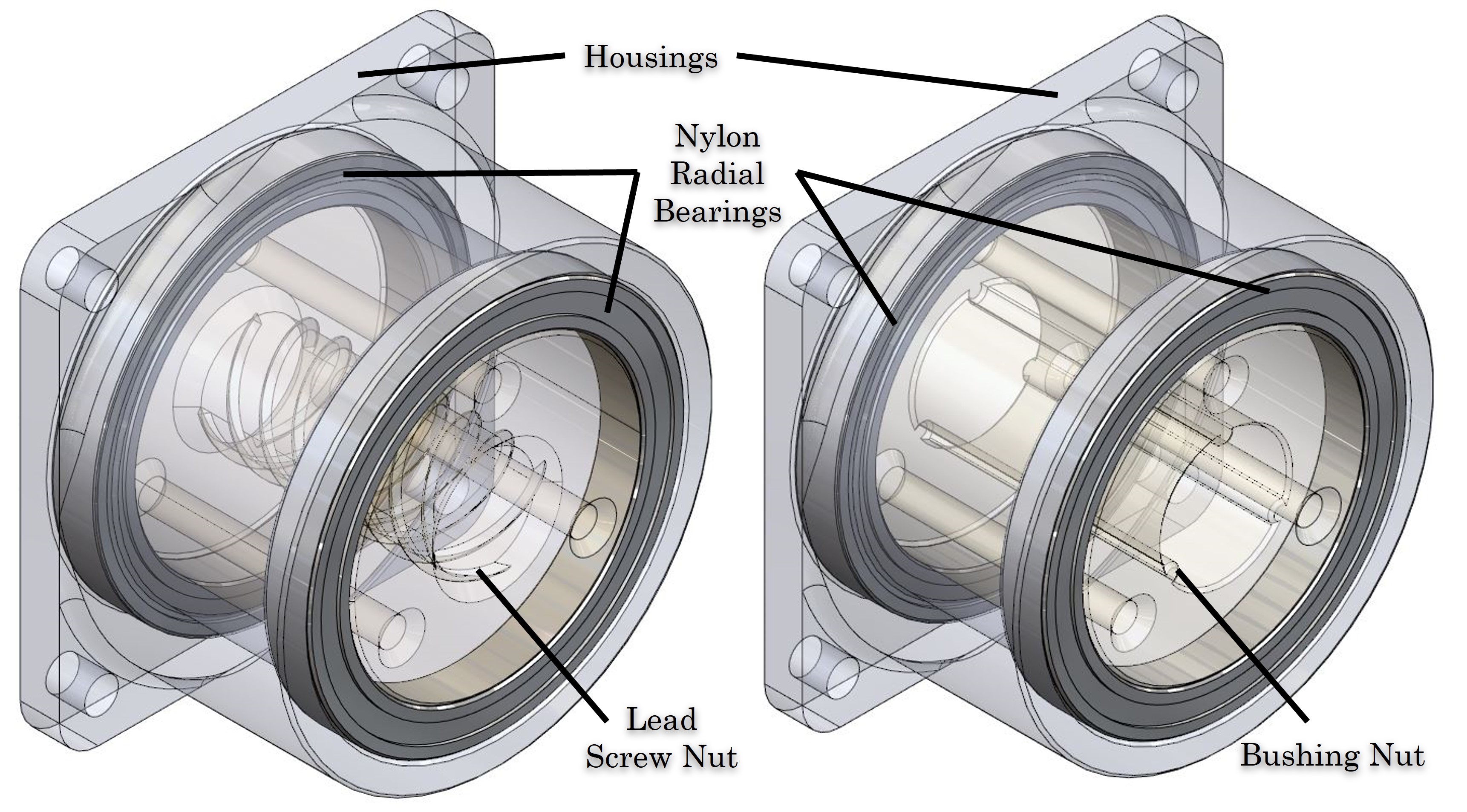}
    \caption{(Left) Lead screw nut assembly. (Right) Lead bushing nut assembly.}
    \label{fig:LS_B}
\end{figure}

Additional to the ball screw/spline design, another similar design based on a lead screw/bushing (LSB) actuator was also developed. Figure \ref{fig:LSBparts} shows the components of the lead screw/busing design, which consisted of a 4-start 20$\times$20mm lead screw nut (DST-J350RM4540DS20$\times$20, Igus, Germany), a custom-designed bushing nut with 3$\times$5/32in bushing fins equally distributed in the $\phi$20mm center hole, and a shaft with bushing groove and screw thread matching with two nuts above respectively. The two nuts were all Igus dryspin\textregistered~ iglide\textregistered~ J material, specifically, the screw nut was an in-stock product with additional modification, while the bushing was a custom 3D printing part using iglide material. In Figure \ref{fig:LS_B} the two nuts assemblies are presented. Using the same housings and plastic bearing pairs, the two nuts were held and enclosed with housings and bearings. 

\begin{figure}
    \centering
    \includegraphics[width=0.9\linewidth]{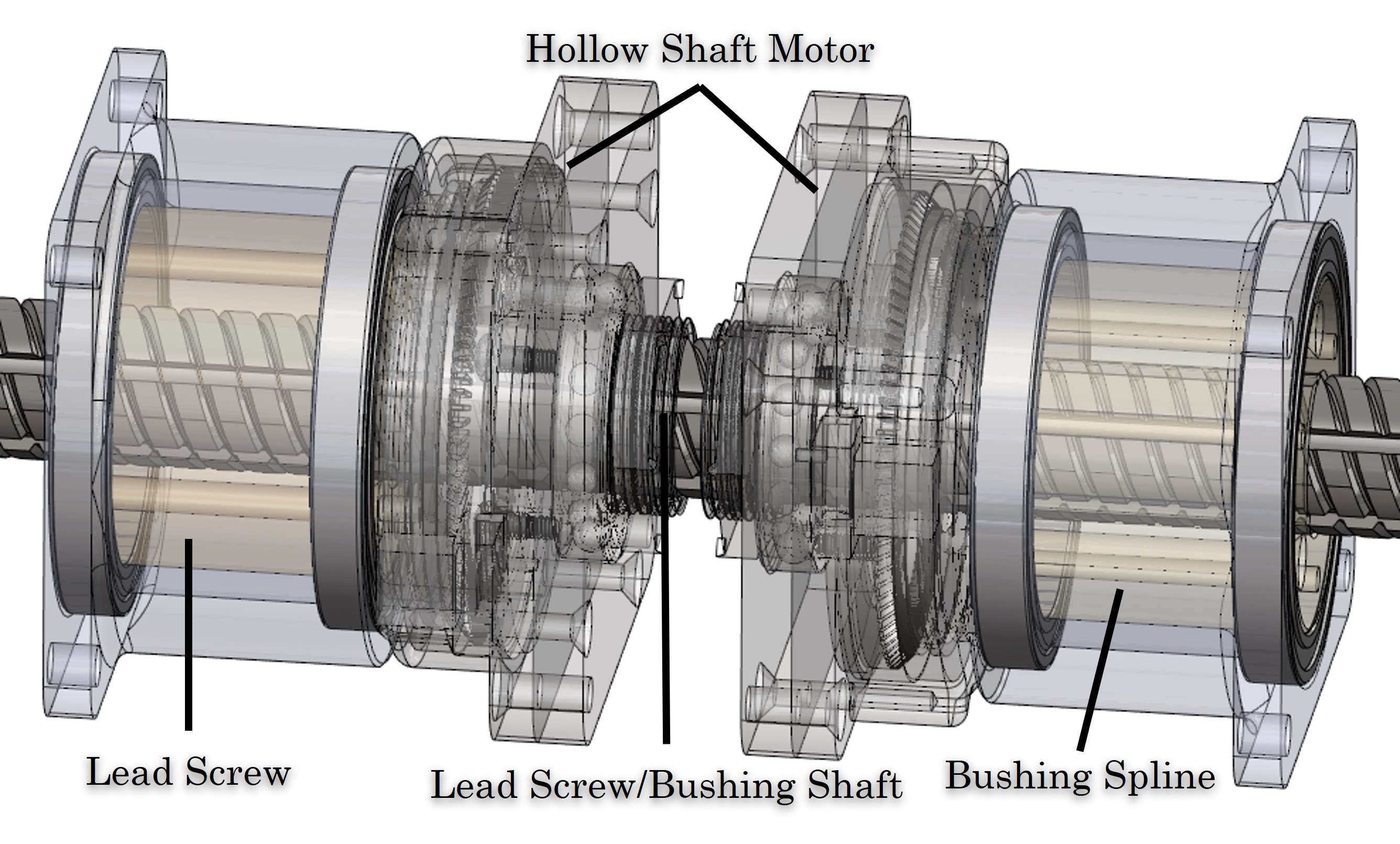}
    \caption{Hollow type of differential drive using lead screw/bushing configuration.}
    \label{fig:LSBdesign}
\end{figure}

Figure \ref{fig:LSBdesign} shows the holly type of differential drive using LSB configuration. Same as the BSS configuration the two hollow shaft motors were fixed with two nuts and the housings were fixed with motors onto the robot, and a grooved shaft was through both nuts and motors directly contacting with screw thread and grooves. The LSB actuator design has fewer parts and is easier to manufacture, also it just is easier to recover from over-driving out of the nut compared to the BSS actuator since there are no free balls inside the nuts. Considering the ball spline and lead bushing were using the same dimension design, these two nuts actuators are exchanged. And for easy manufacturing purposes, we were using a combination of ball screw and lead bushing hybrid (BSLB) actuator in the next section.

\section{System Design}

In the previous section, we introduced two types of differential drive designs, especially the modular design of ball spline and lead bushing nut which are interchangeable. For proof of concept and easy fabrication purposes, we used the combination of ball screw and lead bushing hybrid (BSLB) actuator for the preliminary prototyping. In this section, two-needle drive sub-systems, namely hollow and side-pulley types using BSLB driving systems were presented and discussed.

\subsection{Hollow Type Differential Drive}

\begin{figure}
    \centering
    \includegraphics[width=1\linewidth]{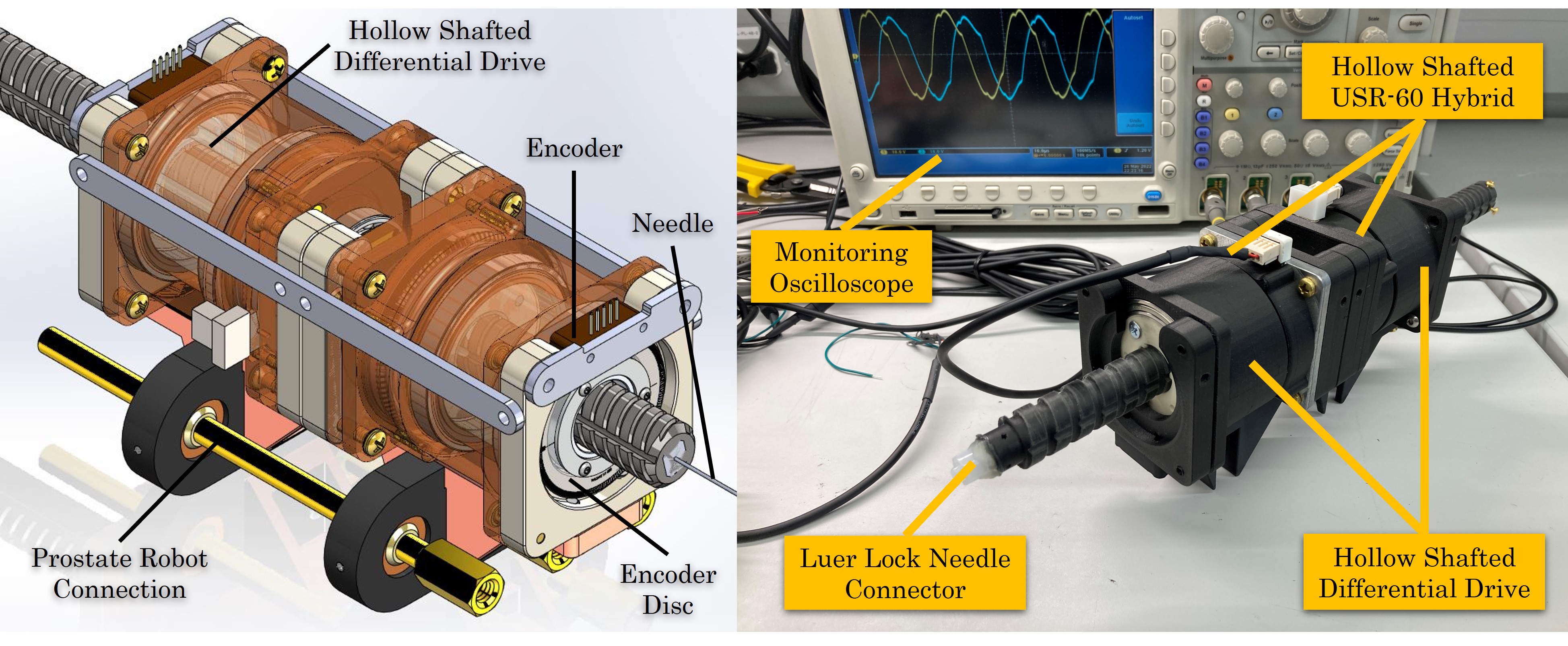}
    \caption{Hollow type differential drive (HDD) subsystem with robot switchable configuration. (Left) Design of hollow differential drive actuator sitting on the prostate robot connection mechanism. (Right) Preliminary prototype of HDD subsystem.}
    \label{fig:HollowDR}
\end{figure}

A hollow-type differential drive (HDD) subsystem with a robot switchable configuration is shown in Figure \ref{fig:HollowDR}. The left figure shows the design of a hollow differential drive actuator sitting on the prostate robot (robot reported in \cite{li2014robotic,li2015development,li2016robotic,wartenberg2018bore,wartenberg2018towards,tavakkolmoghaddam2023passive,tavakkolmoghaddam2023design}) connection mechanism. Two 2-inch encoder discs were fixed on the external side of two nuts, and the matching encoders were fixed onto the support with the robot body. The needle was inserted and fixed into the shaft. A prototype of the HDD is shown in the right figure. Two hybrid HSMs integrated with the original aluminum stage were used in the system to drive both nuts accordingly. A Luer lock needle connector was fixed on the top of the shaft for needle installation, and a monitoring oscilloscope was used to monitor the signal of two HSMs. 

The HDD subsystem shown above is the original design with the most compact configuration. However, because of the limitation and dimensional tolerance of manual fabrication, the performance of HSMs was not very good, which was not capable of driving nuts with friction. What's worse, the rotary speed provided by the motor was slow, measurement of rotary speed yields approximately 75rpm, which was enough for a normal biopsy procedure. However for a later application like CURV steering mentioned in \cite{li2016robotic,wartenberg2018closed,tavakkolmoghaddam2023passive,tavakkolmoghaddam2023design} which required high rotary speed motion, the direct drive from HSMs was not sufficient. An acceleration mechanism integrated differential drive system needs to be developed to fulfill the requirement. 

\subsection{Side-Pulley Type for Prostate Robot}

\begin{figure}
    \centering
    \includegraphics[width=1\linewidth]{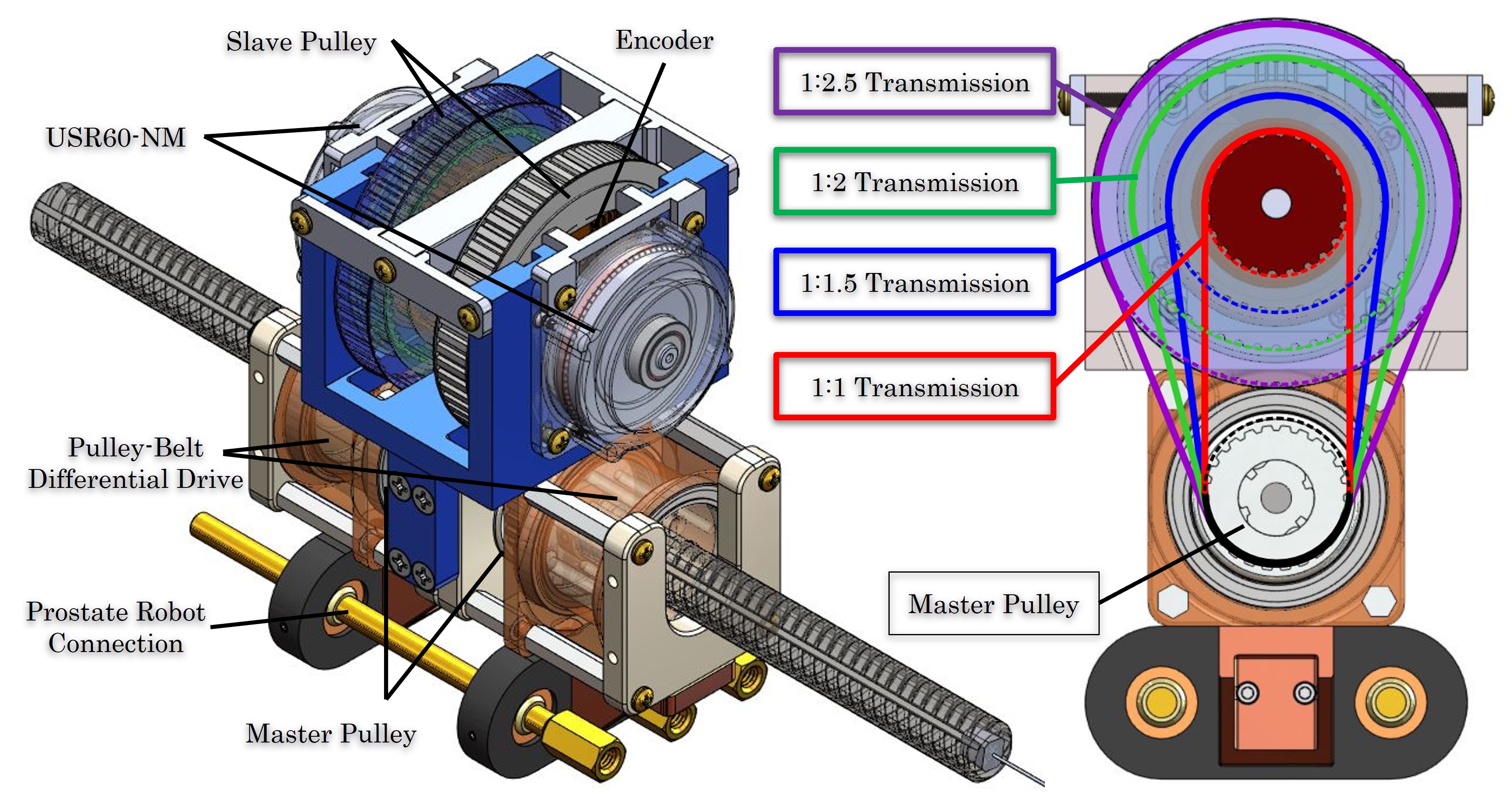}
    \caption{(Left) Design of side-pulley type of differential drive (PDD). (Right) Acceleration transmission configuration based on the combination of pulleys pair.}
    \label{fig:LeadDRdesign}
\end{figure}

To address the need for rotary speed at a high level, a side-pulley type of differential drive (PDD) was developed and presented in this study. Figure \ref{fig:LeadDRdesign} shows the PDD design based on the LSB actuator. Instead of using HSMs for direct drive, two modified $\Phi$1.528in, 24 teeth nylon pulleys with $\phi$22mm center hole were used as master pulley connected to each nut of the LSB actuator, where the whole system was seated onto the prostate robot connection mechanism. A holder with a chamfered hole was located on the top of the LSB actuator with multiple slave pulley pairs, and two USR60-NM were driving the slave pulleys. The right figure shows the multiple-speed transmission generated by different pulley pairs, and four different transmission ratios were shown. Detailed configuration can be found in Table \ref{tbl:trans}. 

\begin{table}[ht]
    \centering
    \caption{Pulleys pair transmission configuration. * Slave pulley 5 is not shown in the figure and is oversized in the current design. Data measured without shaft.}
    \resizebox{\linewidth}{!}{
    \begin{tabular}{cccccc}
    \hline
    Pulley Name     & OD (in) & \begin{tabular}[c]{@{}c@{}}Teeth\\ Number\end{tabular} & \begin{tabular}[c]{@{}c@{}}Transmission\\ Ratio\end{tabular} & \begin{tabular}[c]{@{}c@{}}Rated Speed\\ (RPM)\end{tabular} & \begin{tabular}[c]{@{}c@{}}Real Speed\\ (RPM)\end{tabular} \\ \hline\hline
    Master Pulley   & 1.528   & 24    & -   & 150  & 75     \\ \hline
    Slave Pulley 1  & 3.8     & 60    & 1:2.5 & 375  & 187.5  \\ \hline
    Slave Pulley 2  & 3.056   & 48    & 1:2   & 300  & 150    \\ \hline
    Slave Pulley 3  & 2.292   & 36    & 1:1.5 & 225  & 112.5  \\ \hline
    Slave Pulley 4  & 1.528   & 24    & 1:1   & 150  & 75     \\ \hline
    Slave Pulley 5* & 4.584   & 72    & 1:3   & 450  & 225     \\ \hline
    \end{tabular}
    }
    \label{tbl:trans}
\end{table}

\begin{figure}
    \centering
    \includegraphics[width=1\linewidth]{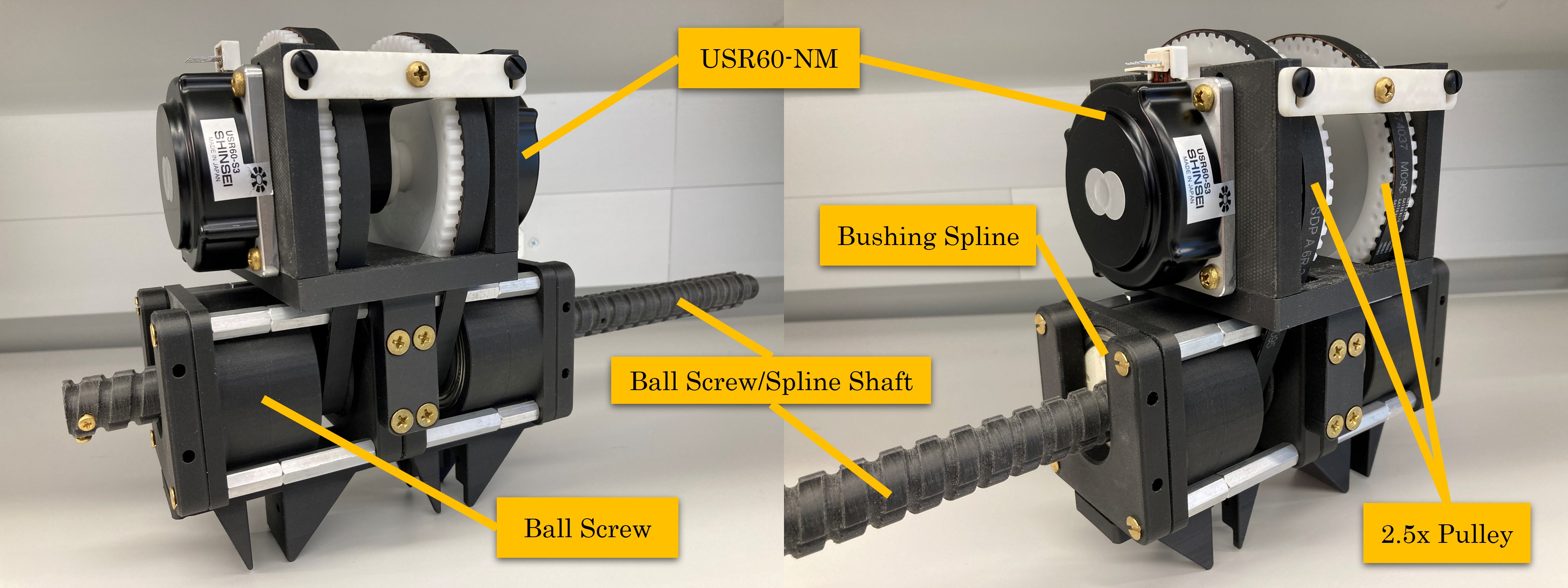}
    \caption{Prototype of PDD system with hybrid BSLB actuator.}
    \label{fig:LeadDR}
\end{figure}

The prototype of the PDD system is shown in Figure \ref{fig:LeadDR}. For proof of concept and easy fabrication purposes, we used a combination of BSLB actuators same as the hollow type. Two 3D-printed housings enclosed the ball screw nut and lead bushing nut and hold by two paired plastic bearings. Two modified nylon master pulleys were fixed on the inner surface of the ball screw nut and lead bushing nut respectively and paired with slave pulley 1 configuration of a 1:2.5 transmission ratio. A Timing XL belt was connected with the master and slave pulleys and held by the top motor assembly. Two USR60-NM motors were connected with the slave pulleys on one side and the center mutual side, there was a small housing holding two small bearings to balance the pulleys. Motor-slave pulleys assembly was designed to be adjustable on vertical location to load and release tension to the belts. Three ball screw/spline shaft units with 120mm length were connected making the main shaft to be a 360mm length. A male Luer lock was fixed on the top of the shaft for needle installation. In this study, the main validation and testing were using this PDD setup as the main switchable subsystem for prostate robot needle driving.
 
\subsection{Side-Pulley Type for NeuroAblation Robot}

\begin{figure}
    \centering
    \includegraphics[width=1\linewidth]{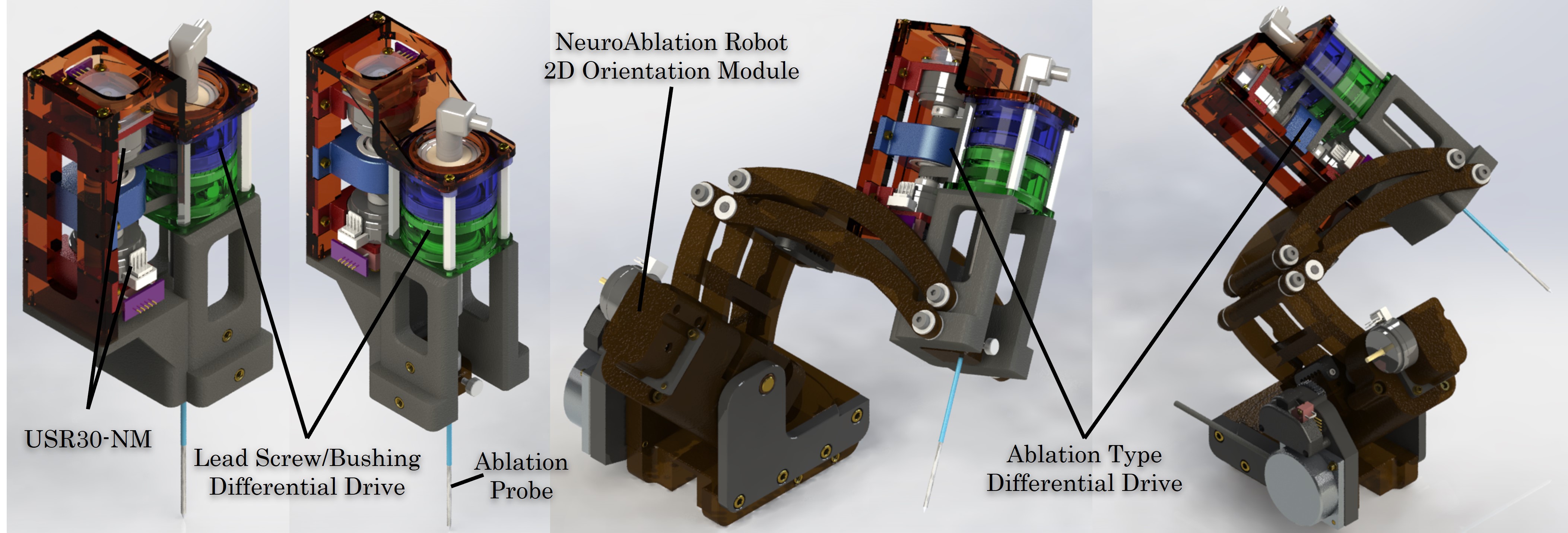}
    \caption{A minimized LSB type PDD system integrated with the current NeuroAblation robot 2D orientation module.}
    \label{fig:DRneurorobot}
\end{figure}

A conception of a smaller-size PDD system was also designed for our previous MRI-compatible NeuroAblation robot \cite{nycz2017mechanical} which is shown in Figure \ref{fig:DRneurorobot}. The advantages of using HCM are significant, and another solution of using a differential drive system to control the motion of the needle-based probe is presented here. By using small USM USR30-NM motors, a differential drive system was integrated into the NeuroAblation robot to replace the needle insertion and rotation motion. More studies about brain tumor ablation procedures can be found in our previous work \cite{campwala2021predicting,szewczyk2022happens,gandomi2019thermo,gandomi2020modeling,tavakkolmoghaddam2021neuroplan,jiang2024icap,zhao2024deep,zhao2024development}. Similar to the PDD top motors holding mechanism, two USR30-NM were fixed onto the robot arm and connected with pulleys respectively. Belts connected the master and slave pulleys and transmitted motion to the master pulleys. To minimize the volume, master pulley teeth were manufactured on the lateral wall of the lead screw and bushing nuts directly. An ablation probe integrated with screw thread and bushing groove on the body was inserted into the minimized LSB nuts center hole and controlled by the LSB-type PDD system. The whole subsystem was designed based on the current NeuroAblation robot 2D orientation module, which was also compatible with the current probe driver module with switchable ability. This subsystem should be developed further in future work.

\section{System Validation}

\begin{figure}
    \centering
    \includegraphics[width=0.9\linewidth]{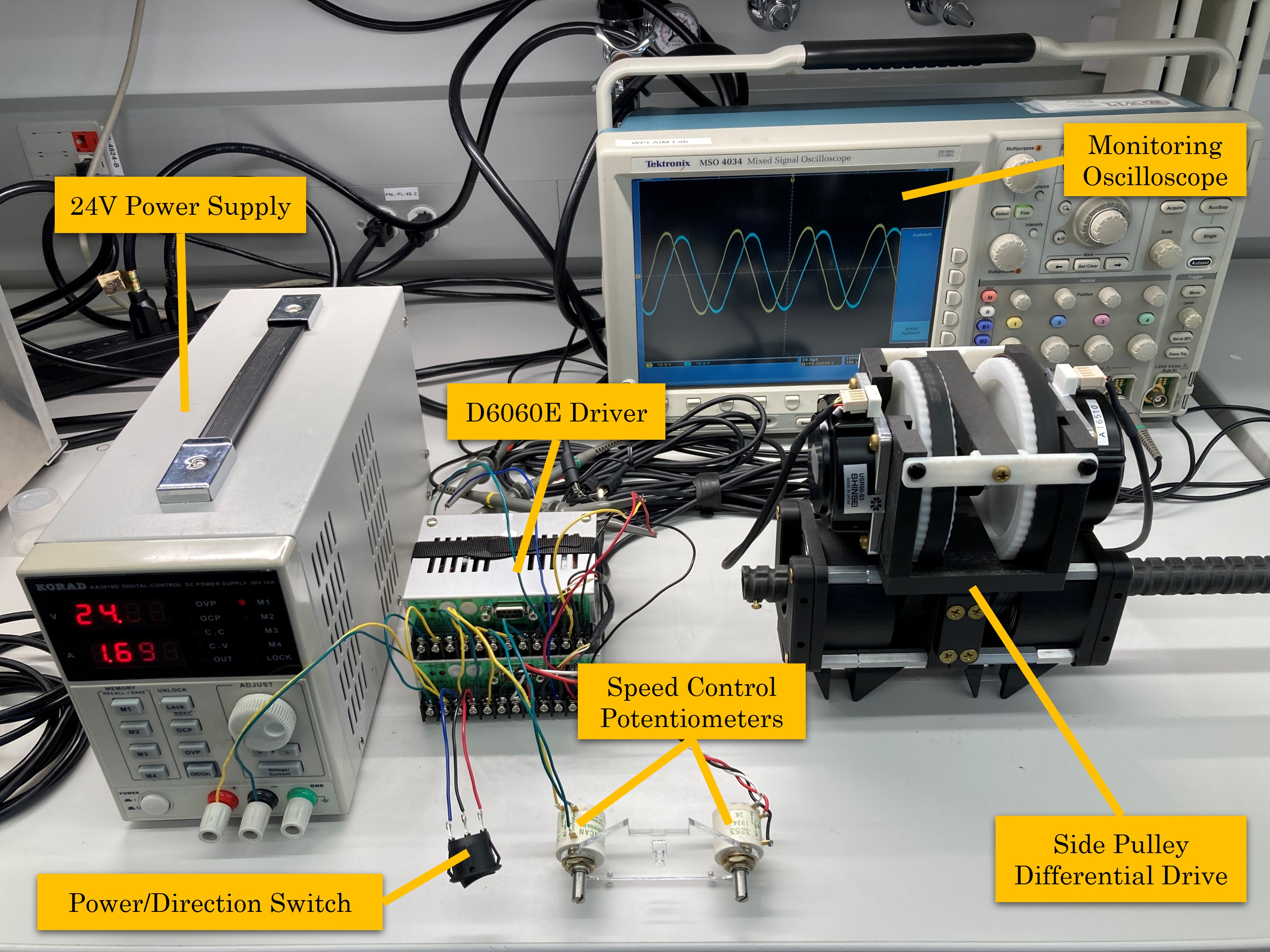}
    \caption{Basic functionality testing setup for hollow shaft motors and differential drive.}
    \label{fig:TestingSetup}
\end{figure}

A series of validations are discussed including the HCM validation, mechanical functionality, motion control, and accuracy validation. The basic functionality testing setup can be found in Figure \ref{fig:TestingSetup}. Two D6060E drivers from Shinsei were used to control the USR60-NM motors on the PDD system, which were powered by a 24V power supply. A 3-position switch was used to trigger the power and clockwise (CW) and counterclockwise (CCW) rotary direction control, and two potentiometers were connected to each driver for speed control. The output signals of both drivers were connected to each motor accordingly and also connected to an oscilloscope for monitoring. Note that considering the rotation motion of the needle required the same direction and same speed from both motors, for the initial functionality test we calibrated the speed synchronization visually by the speed control potentiometers. 

Results show that the custom-made HCM was working functionally with a rotary speed of 75rpm at zero load. However, the torque generated by the 3D printed rotor pairs was low, which struggled to rotate the HDD actuator. This was because the surface roughness of the new custom-made rotor was not modified for the new rotor. For the PDD system, the two motors were able to drive the 1:2.5 acceleration configuration, with nuts rotation at 168rpm with shaft. Some deduction of the speed came from the non-concentric of the shaft and two rotary nuts misalignments.

\subsection{Robot Setup}

\begin{figure}
    \centering
    \includegraphics[width=0.9\linewidth]{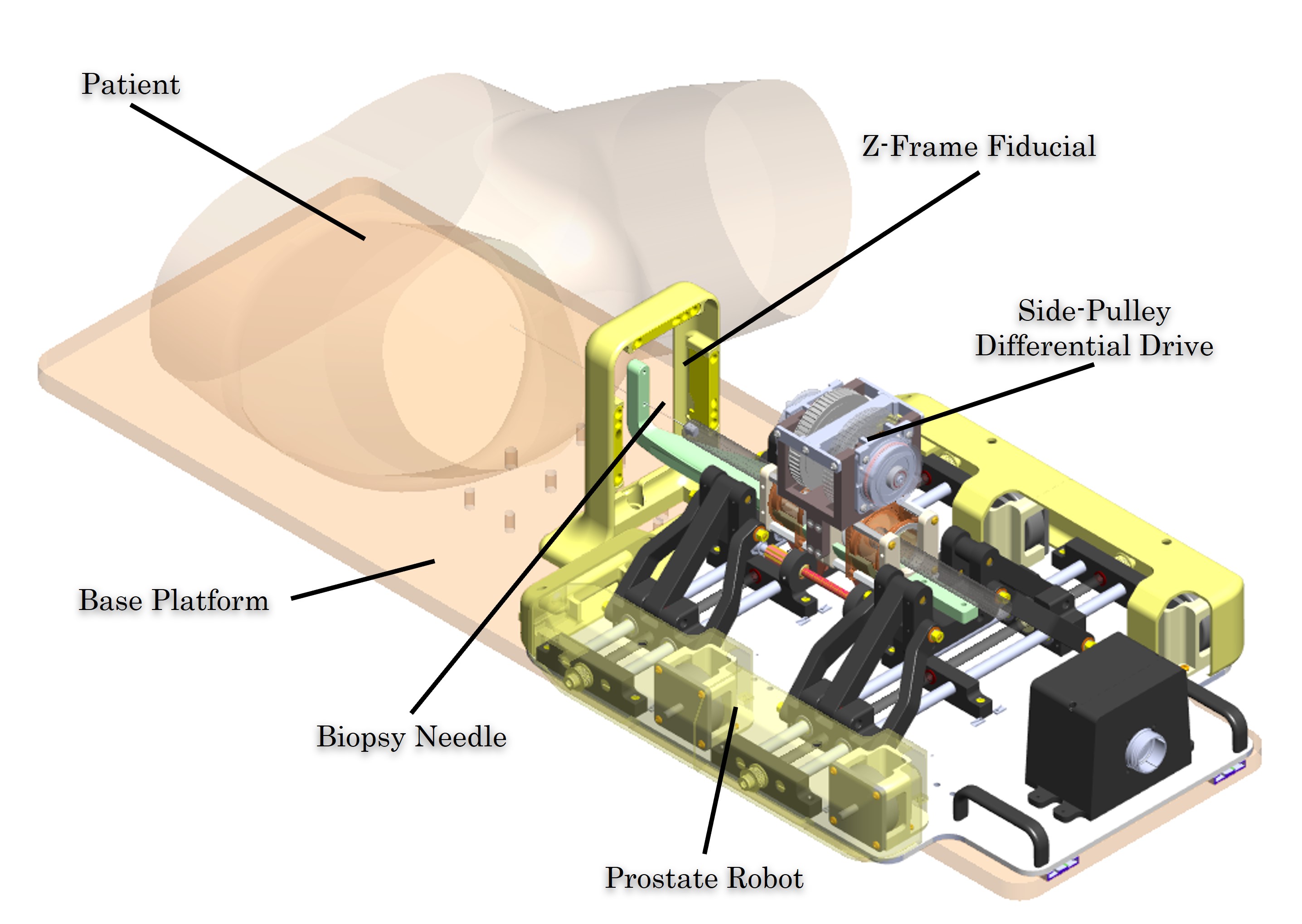}
    \caption{Pulley differential drive equipped onto the prostate robot and deployed on the base platform with dummy patient.}
    \label{fig:AccPulley}
\end{figure}

\begin{figure}
    \centering
    \includegraphics[width=1\linewidth]{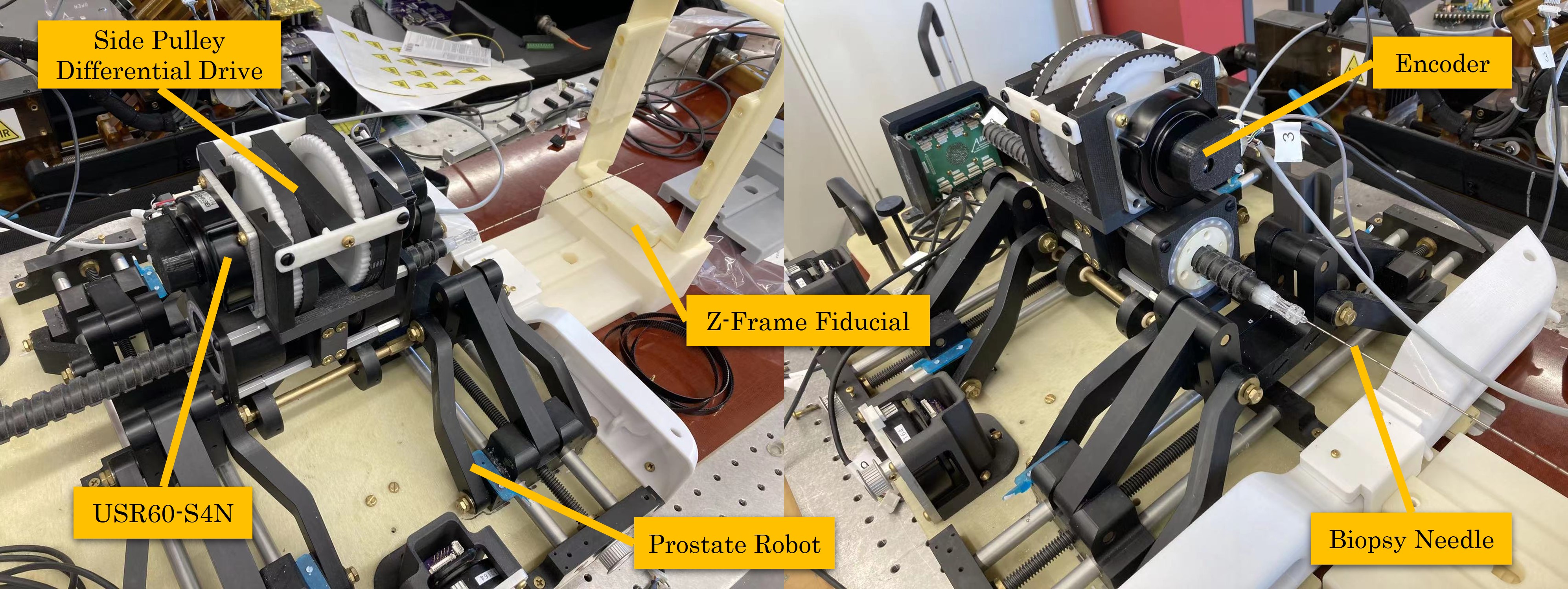}
    \caption{Differential drive equipped onto prostate robot with a needle attached.}
    \label{fig:DRrobot1}
\end{figure}

A PDD system integrated prostate robot design is shown in Figure \ref{fig:AccPulley}. The PDD subsystem was attached to the 4-DOF prostate robot top with an L-shape needle guide connected, the biopsy needle was connected on the initial top of the shaft and through the guide. The patient, Z-frame fiducial, and robot system were all fixed onto the base platform, and all the components shown in the figure were located on the MRI table. A PDD-integrated prostate robot system can be found in Figure \ref{fig:DRrobot1}, which follows the same configuration mentioned above. Note that the motors used in the robot were USR60-S4N, which had a back shaft at the back of the motor, so the encoder (EM1-0-1250-N, US Digital) and disc can be attached to the back of the motors. 

\begin{figure}
    \centering
    \includegraphics[width=0.7\linewidth]{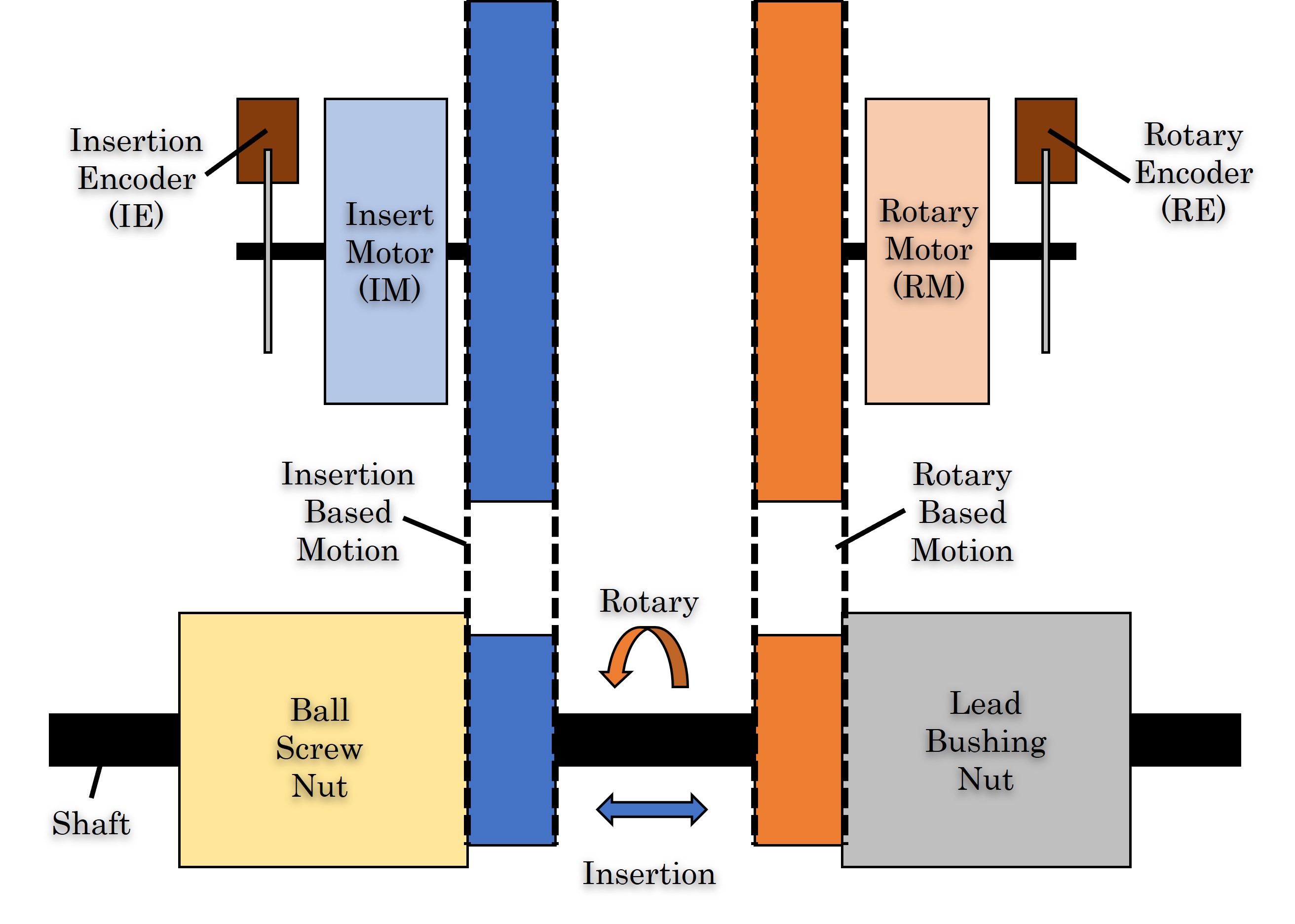}
    \caption{Schematic image of side-pulley type differential drive. The IM/IE system was set to the main actuator of insertion, while the RM/RE was set to the main actuator of the rotary.}
    \label{fig:control}
\end{figure}

The principle of ``differential'' is based on the combination motions of the two nuts or two driving motors specifically. In this study, we only discuss the basic functionality of general BSS system motion, namely linear motion, and rotary motion, and a supplementary argument code was implanted into our prostate robot controlling code both in high-level C++ and engineering UI. The schematic image of PDD is shown in Figure \ref{fig:control}. The shaft rotary and insertion motion was achieved by the combination motions of the ball screw and lead bushing nut, which were driven by two motors respectively. In our current controlling system, each DOF motion was one-to-one mapping with a single motor, and encoders were used to send position feedback to the control box and display it on the UI page. The new differential drive segment controller was using two motors mapping to rotary motion, so when rotary motion was processed, two motors were spinning simultaneously, under this circumstance one of the encoders had to stop reading or sending any position feedback to the control box for compensation.  We labeled the two motors and encoders as insertion motor (IM)/insertion encoder (IE) and rotary motor (RM)/rotary encoder (RE), and a general position feedback equation of the 2-DOF differential drive position algorithm can be found in Equation \ref{equ:IK}, where the insertion and rotary represented the positioning value showing on the UI, and IE and RE described the value reading from each encoder. This equation indicates that the insertion position was always subtracted from the reading value of the rotary encoder. This equation was defined as the general controller pattern, however, in this study considering the goal of proof of concept, a simplified control method was developed. 

\begin{equation}
    \begin{aligned}
        & Insertion = IE - RE\\
        & Rotary = RE\\
    \end{aligned}
    \label{equ:IK}
\end{equation}

After labeling the motors and encoders, we set the primary motor/encoder for each motion. In this preliminary study, we developed a bang-bang controller for updating the motors' enable/disable status and the encoders updating regulation to perform insertion, rotary, and mixed motion for needle driving purposes. For insertion motion, it was the same one-to-one mapping as the current controller, however, for rotary motion, the IE reading value will affect the current position value of insertion motion. By forcing the IM to follow the RM and stop counting IE, the interference from IE was reduced. We first set up a trigger from the UI, which was the Needle Rotation Enable button to trigger an additional controller mode with IM following RM and IE stop counting. When the Rotation Enable button was disabled, the robot set back to the original controller with normal motion control. This method solves the issue of multi-mapping and can achieve the basic motion function of the new PDD system. 

\subsection{Accuracy Validation}

\begin{figure}
    \centering
    \includegraphics[width=1\linewidth]{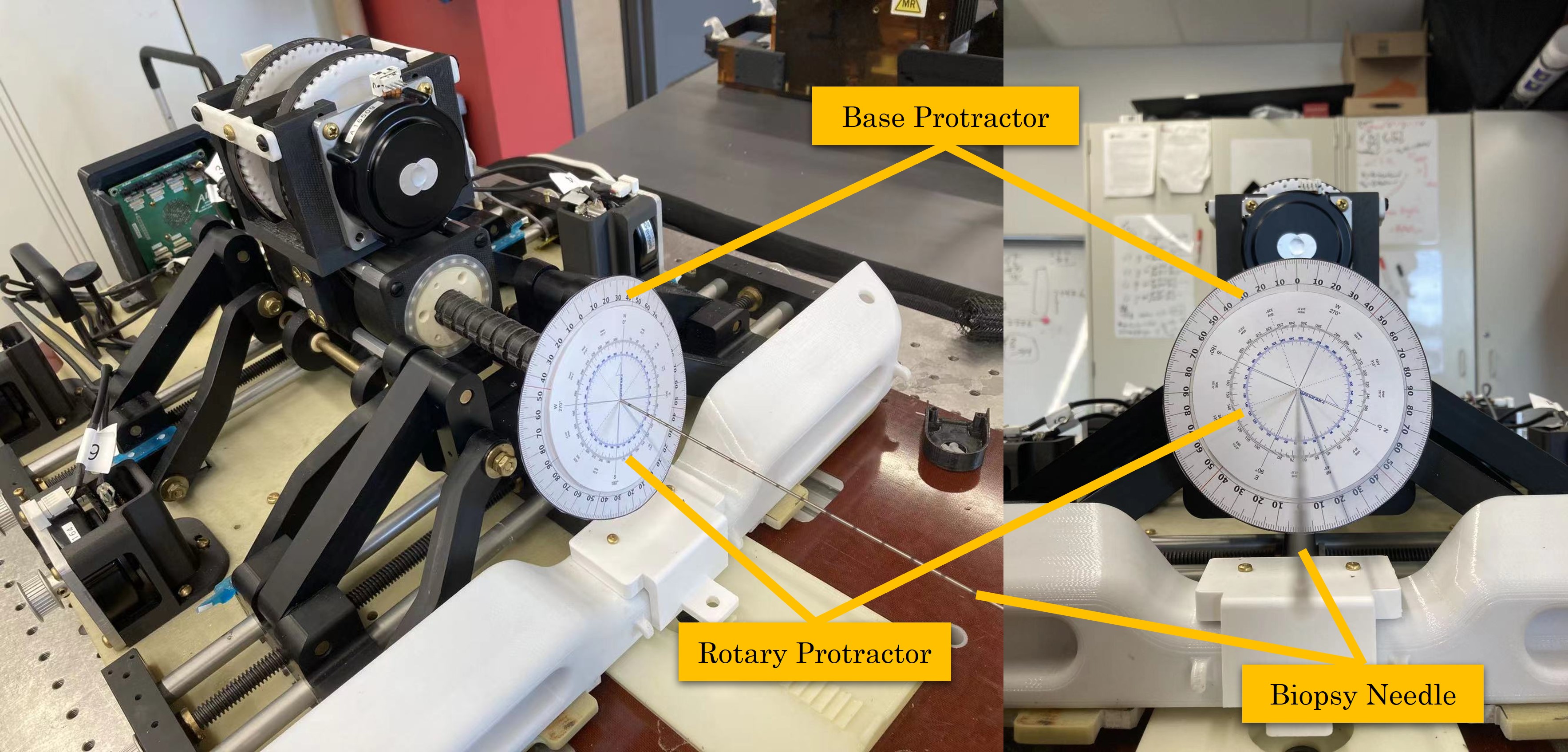}
    \caption{Needle rotation validation setup.}
    \label{fig:DRrobotMeasure}
\end{figure}

After validating the basic function and proving the new controller, system accuracy validation is required. In this section, a preliminary accuracy validation experiment and results are discussed. For proof of concept purposes, the PDD system was validated using a caliper and protractor for insertion and rotary accuracy test. For the insertion accuracy experiment, we chose the front gantry surface as the baseline and the top surface of the Luer lock as the testing surface. By using the control system to the target point enable the insertion motion, and read the displacement of the baseline and target surface with five trials for each target distance. Figure \ref{fig:DRrobotMeasure} shows the needle rotation validation setup, which was using the difference of two protractors as experimental data. The two protractors were located on the front side of the shaft, where the base protractor was fixed on the front gantry surface with a through hole in the center for the shaft being through, while the rotary protractor was fixed onto the needle side of the Luer connection to prevent any contact interference generated from the two protractors.

The rotary speed was measured, which reached up to approximately 168rpm, and Table \ref{tbl:insert} summarizes the insertion axis ($I_i$) accuracy test results. The highest insertion means error ($\mu_{I_i}$) is -1.9mm and the highest insertion standard deviation ($\sigma_{I_i}$) is 3.43mm. Results indicate that the larger the distance insertion applied, the larger the error it performed. This was because of the manually manufactured dimensional error and tolerance which accumulated that error with long-distance of operation. Table \ref{tbl:rotary} summarizes the needle driver rotation axis ($R_i$) accuracy test results. The largest rotation mean error ($\mu_{R_i}$) is 1.45$^\circ$, and the largest rotation standard deviation ($\sigma_{R_i}$) is 1.789$^\circ$. Repeatability in this axis was expected since the needle will require over 360$^\circ$ rotation, as well as the requirement of the CURV steering method for needle targeting compensation. Note that the error of rotary accuracy was only performed within the two-cycle (at most 720$^\circ$) range, where the interference to the IM and IE was small and can be neglected. A further experiment of multiple cycle rotation for accumulation error observed a large error, namely every seven circles of needle rotation (2520$^\circ$) there was approximately 2.1mm insertion accumulation feed-in error displacement. This was because of the speed misalignment between the RM and IM, and this tiny amount of difference in speed caused a tiny spiral motion instead of pure rotary motion. With the accumulation of these spiral motions after a huge amount of rotation motion, the insertion error will be large. To solve this issue, a delicate motor speed controller like a PID compensation controller is desired for further study.

\begin{table}[ht!]
    \centering
    \caption{Insertion Accuracy (n=5). Unit - mm}
    \begin{tabular}{|c|c|c|c|}
    \hline
    Target Points & Target Location $I_i$ & $\mu_{I_i}$    & $\sigma_{I_i}$   \\ \hline
    I1            & 122               & 0.1  & 2.83 \\ \hline
    I2            & 164.3             & -1.9 & 3.43 \\ \hline
    I3            & 45.3              & -0.94  & 1.16 \\ \hline
    I4            & 162.5             & 1.08  & 3.37 \\ \hline
    I5            & 8.3               & 1.1   & 0.74 \\ \hline
    \end{tabular}
    \label{tbl:insert}
\end{table}

\begin{table}[ht!]
    \centering
    \caption{Rotary Accuracy (n=5). Unit - degrees}
    \begin{tabular}{|c|c|c|c|}
    \hline
    Target Points & Target Location $R_i$ & $\mu_{R_i}$    & $\sigma_{R_i}$  \\ \hline
    R1            & 182.525            & -1.325 & 1.304  \\ \hline
    R2            & 95.4               & -1.06  & 1.549 \\ \hline
    R3            & 356.75             & 1.45   & 1.789 \\ \hline
    R4            & 142.51             & -0.79  & 1.221 \\ \hline
    R5            & 8.825              & 0.835  & 0.422   \\ \hline
    \end{tabular}
    \label{tbl:rotary}
\end{table}

\section{Conclusion and Disscussion}

In this work, we introduced two mechanical differential drive designs: the ball screw/spline design and the lead screw/bushing design. We also explored two types of differential drives, namely the hollow type and the side-pulley type. Additionally, we discussed the MRI-compatible hollow ultrasonic motor (USM) design. Finally, we presented the design and mechanical validation of a 2-degree-of-freedom (2-DOF) preliminary differential drive system intended as an add-on module for prostate robot needle-based biopsy procedures. This preliminary system was developed to create an initial prototype based on a hollow shaft (cylindrical) design and to provide enhanced solutions for needle insertion and rotary manipulation. The system was initially constructed using a low-resolution rapid-prototyping method for proof-of-concept purposes.

The feasibility of using a differential drive was confirmed, and a rapid method for controlling the dual motors and encoders was developed for preliminary validation. The basic functionality of rotary and insertion movements was achieved through a user interface (UI) trigger for mode changes. However, a more refined controller, such as a PID compensation controller, is necessary for precise control.

The custom-made hollow-type hybrid USM demonstrated good rotary performance, reaching a speed of 75 rpm. This performance may have been limited by the unoptimized contact surface of the rotor, which was 3D-printed, and by concentricity issues caused by manual assembly defects. Although this speed was insufficient for high-speed needle rotary applications, the side-pulley type differential increased the speed to 168 rpm, suitable for driving needle rotary motion. The system's accuracy was also assessed, with a mean error and maximum standard deviation of $\mu_{I_i}$=1.9mm, $\sigma_{I_i}$=3.43mm in insertion motion, and $\mu_{R_i}$=1.45$^\circ$, $\sigma_{R_i}$=1.789$^\circ$ in rotary motion. Based on these error values, the proof-of-concept differential drive design shows promise for further development.

Future work should focus on advancing the controller development and fabricating a new generation of differential drives with higher resolution and tighter tolerances. Additionally, improving the hardware for dual-motor speed and direction control is crucial. Given that the symmetry of dual-motor systems significantly impacts the accuracy of differential drives, future development should include speed control and compensation for opposite direction speeds.

\bibliographystyle{IEEEtran}
\bibliography{Main}

\begin{thebibliography}{10}
\providecommand{\url}[1]{#1}
\csname url@samestyle\endcsname
\providecommand{\newblock}{\relax}
\providecommand{\bibinfo}[2]{#2}
\providecommand{\BIBentrySTDinterwordspacing}{\spaceskip=0pt\relax}
\providecommand{\BIBentryALTinterwordstretchfactor}{4}
\providecommand{\BIBentryALTinterwordspacing}{\spaceskip=\fontdimen2\font plus
\BIBentryALTinterwordstretchfactor\fontdimen3\font minus \fontdimen4\font\relax}
\providecommand{\BIBforeignlanguage}[2]{{%
\expandafter\ifx\csname l@#1\endcsname\relax
\typeout{** WARNING: IEEEtran.bst: No hyphenation pattern has been}%
\typeout{** loaded for the language `#1'. Using the pattern for}%
\typeout{** the default language instead.}%
\else
\language=\csname l@#1\endcsname
\fi
#2}}
\providecommand{\BIBdecl}{\relax}
\BIBdecl

\bibitem{chatterjee2019revisiting}
A.~Chatterjee, A.~J. Gallan, D.~He, X.~Fan, D.~Mustafi, A.~Yousuf, T.~Antic, G.~S. Karczmar, and A.~Oto, ``Revisiting quantitative multi-parametric mri of benign prostatic hyperplasia and its differentiation from transition zone cancer,'' \emph{Abdominal Radiology}, vol.~44, no.~6, pp. 2233--2243, 2019.

\bibitem{wartenberg2018towards}
M.~Wartenberg, ``Towards hands-on cooperative control for closed-loop mri-guided targeted prostate biopsy,'' Ph.D. dissertation, The Worcester Polytechnic Institute, 2018.

\bibitem{carvalho2020demonstration}
P.~A. Carvalho, C.~J. Nycz, K.~Y. Gandomi, and G.~S. Fischer, ``Demonstration and experimental validation of plastic-encased resonant ultrasonic piezoelectric actuator for magnetic resonance imaging-guided surgical robots,'' \emph{Journal of engineering and science in medical diagnostics and therapy}, vol.~3, no.~1, p. 011002, 2020.

\bibitem{carvalho2020study}
P.~A. Carvalho, H.~Tang, P.~Razavi, K.~Pooladvand, W.~C. Castro, K.~Y. Gandomi, Z.~Zhao, C.~J. Nycz, C.~Furlong, and G.~S. Fischer, ``Study of mri compatible piezoelectric motors by finite element modeling and high-speed digital holography,'' in \emph{Advancements in Optical Methods \& Digital Image Correlation in Experimental Mechanics, Volume 3}.\hskip 1em plus 0.5em minus 0.4em\relax Springer, 2020, pp. 105--112.

\bibitem{zhao2021preliminary}
Z.~Zhao, P.~Carvalho, H.~Tang, K.~Pooladvand, K.~Gandomi, C.~Nycz, C.~Furlong, and G.~Fischer, ``Preliminary characterization of a plastic piezoelectric motor stator using high-speed digital holographic interferometry,'' in \emph{Advancement of Optical Methods \& Digital Image Correlation in Experimental Mechanics}.\hskip 1em plus 0.5em minus 0.4em\relax Springer, 2021, pp. 89--93.

\bibitem{zhao2023preliminary}
Z.~Zhao, Y.~Wang, D.~Ruiz-Cadalso, H.~Zheng, C.~Bales, F.~Tavakkolmoghaddam, Y.~Jiang, A.~Salerni, C.~Furlong, and G.~Fischer, ``Preliminary characterization of a hollow cylindrical ultrasonic motor by finite element modeling and digital holographic interferometry,'' in \emph{Society for Experimental Mechanics Annual Conference and Exposition}.\hskip 1em plus 0.5em minus 0.4em\relax Springer, 2023, pp. 9--15.

\bibitem{zhao2024study}
Z.~Zhao, H.~Tang, P.~Carvalho, C.~Furlong, and G.~S. Fischer, ``Study of mri-compatible notched plastic ultrasonic stator with fem simulation and holography validation,'' \emph{arXiv preprint arXiv:2408.08528}, 2024.

\bibitem{zhao2024design}
Z.~Zhao, C.~Bales, and G.~Fischer, ``Design and characterization of mri-compatible plastic ultrasonic motor,'' \emph{arXiv preprint arXiv:2409.04006}, 2024.

\bibitem{zhao2024characterization}
Z.~Zhao, Y.~Wang, C.~Bales, D.~Ruiz-Cadalso, H.~Zheng, C.~Furlong-Vazquez, and G.~Fischer, ``Characterization and design of a hollow cylindrical ultrasonic motor,'' \emph{arXiv preprint arXiv:2409.07690}, 2024.

\bibitem{li2014robotic}
G.~Li, H.~Su, G.~A. Cole, W.~Shang, K.~Harrington, A.~Camilo, J.~G. Pilitsis, and G.~S. Fischer, ``Robotic system for mri-guided stereotactic neurosurgery,'' \emph{IEEE transactions on biomedical engineering}, vol.~62, no.~4, pp. 1077--1088, 2014.

\bibitem{li2015development}
M.~Li, B.~Gonenc, K.~Kim, W.~Shang, and I.~Iordachita, ``Development of an mri-compatible needle driver for in-bore prostate biopsy,'' in \emph{2015 International Conference on Advanced Robotics (ICAR)}.\hskip 1em plus 0.5em minus 0.4em\relax IEEE, 2015, pp. 130--136.

\bibitem{li2016robotic}
G.~Li, ``Robotic system development for precision mri-guided needle-based interventions,'' Ph.D. dissertation, The Worcester Polytechnic Institute, 2016.

\bibitem{wartenberg2018bore}
M.~Wartenberg, K.~Gandomi, P.~Carvalho, J.~Schornak, N.~Patel, I.~Iordachita, C.~Tempany, N.~Hata, J.~Tokuda, and G.~S. Fischer, ``In-bore experimental validation of active compensation and membrane puncture detection for targeted mri-guided robotic prostate biopsy,'' in \emph{International Symposium on Experimental Robotics}.\hskip 1em plus 0.5em minus 0.4em\relax Springer, 2018, pp. 34--44.

\bibitem{tavakkolmoghaddam2023passive}
F.~Tavakkolmoghaddam, Y.~Wang, C.~Bales, Y.~Jiang, C.~Nycz, Z.~Zhao, and G.~Fischer, ``Passive model-based error compensation for beveled-tip needle deflection,'' in \emph{2023 International Symposium on Medical Robotics (ISMR)}.\hskip 1em plus 0.5em minus 0.4em\relax IEEE, 2023, pp. 1--7.

\bibitem{tavakkolmoghaddam2023design}
F.~Tavakkolmoghaddam, C.~Bales, Y.~Wang, Z.~Zhao, and G.~S. Fischer, ``Design and evaluation of bidirectional continuous rotation and variable curvature needle steering algorithm,'' in \emph{2023 IEEE/RSJ International Conference on Intelligent Robots and Systems (IROS)}.\hskip 1em plus 0.5em minus 0.4em\relax IEEE, 2023, pp. 10\,315--10\,322.

\bibitem{wartenberg2018closed}
M.~Wartenberg, J.~Schornak, K.~Gandomi, P.~Carvalho, C.~Nycz, N.~Patel, I.~Iordachita, C.~Tempany, N.~Hata, J.~Tokuda \emph{et~al.}, ``Closed-loop active compensation for needle deflection and target shift during cooperatively controlled robotic needle insertion,'' \emph{Annals of biomedical engineering}, vol.~46, no.~10, pp. 1582--1594, 2018.

\bibitem{nycz2017mechanical}
C.~J. Nycz, R.~Gondokaryono, P.~Carvalho, N.~Patel, M.~Wartenberg, J.~G. Pilitsis, and G.~S. Fischer, ``Mechanical validation of an mri compatible stereotactic neurosurgery robot in preparation for pre-clinical trials,'' in \emph{2017 IEEE/RSJ International Conference on Intelligent Robots and Systems (IROS)}.\hskip 1em plus 0.5em minus 0.4em\relax IEEE, 2017, pp. 1677--1684.

\bibitem{campwala2021predicting}
Z.~Campwala, B.~Szewczyk, T.~Maietta, R.~Trowbridge, M.~Tarasek, C.~Bhushan, E.~Fiveland, G.~Ghoshal, T.~Heffter, K.~Gandomi \emph{et~al.}, ``Predicting ablation zones with multislice volumetric 2-d magnetic resonance thermal imaging,'' \emph{International Journal of Hyperthermia}, vol.~38, no.~1, pp. 907--915, 2021.

\bibitem{szewczyk2022happens}
B.~Szewczyk, M.~Tarasek, Z.~Campwala, R.~Trowbridge, Z.~Zhao, P.~M. Johansen, Z.~Olmsted, C.~Bhushan, E.~Fiveland, G.~Ghoshal \emph{et~al.}, ``What happens to brain outside the thermal ablation zones? an assessment of needle-based therapeutic ultrasound in survival swine,'' \emph{International Journal of Hyperthermia}, vol.~39, no.~1, pp. 1283--1293, 2022.

\bibitem{gandomi2019thermo}
K.~Gandomi, P.~Carvalho, Z.~Zhao, C.~Nycz, E.~Burdette, and G.~Fischer, ``Thermo-acoustic simulation of a piezoelectric transducer for interstitial thermal ablation with mrti based validation,'' in \emph{Comsol Conference}, 2019.

\bibitem{gandomi2020modeling}
K.~Y. Gandomi, P.~A. Carvalho, M.~Tarasek, E.~W. Fiveland, C.~Bhushan, E.~Williams, P.~Neubauer, Z.~Zhao, J.~Pilitsis, D.~Yeo \emph{et~al.}, ``Modeling of interstitial ultrasound ablation for continuous applicator rotation with mr validation,'' \emph{IEEE Transactions on Biomedical Engineering}, vol.~68, no.~6, pp. 1838--1846, 2020.

\bibitem{tavakkolmoghaddam2021neuroplan}
F.~Tavakkolmoghaddam, D.~K. Rajamani, B.~Szewczyk, Z.~Zhao, K.~Gandomi, S.~C. Sekhar, J.~Pilitsis, C.~Nycz, and G.~Fischer, ``Neuroplan: A surgical planning toolkit for an mri-compatible stereotactic neurosurgery robot,'' in \emph{2021 International Symposium on Medical Robotics (ISMR)}.\hskip 1em plus 0.5em minus 0.4em\relax IEEE, 2021, pp. 1--7.

\bibitem{jiang2024icap}
Y.~Jiang, Y.~Wang, Z.~Zhao, C.~Bales, K.~Gandomi, C.~J. Nycz, and G.~S. Fischer, ``Icap: Interactive conformal ablation planning toolkit with mr thermometry validation,'' \emph{Authorea Preprints}, 2024.

\bibitem{zhao2024deep}
Z.~Zhao, B.~Szewczyk, M.~Tarasek, C.~Bales, Y.~Wang, M.~Liu, Y.~Jiang, C.~Bhushan, E.~Fiveland, Z.~Campwala \emph{et~al.}, ``Deep brain ultrasound ablation thermal dose modeling with in vivo experimental validation,'' \emph{arXiv preprint arXiv:2409.02395}, 2024.

\bibitem{zhao2024development}
Z.~Zhao, Y.~Jiang, C.~Bales, Y.~Wang, and G.~Fischer, ``Development of advanced fem simulation technology for pre-operative surgical planning,'' \emph{arXiv preprint arXiv:2409.03990}, 2024.

\end{thebibliography}

\end{document}